\documentclass[11pt,letterpaper]{article}
\usepackage{naaclhlt2016}
\usepackage{times}
\usepackage{latexsym}
\usepackage{lingmacros}
\usepackage{times}
\usepackage{url}
\usepackage{tikz-dependency}
\usepackage{latexsym}
\usepackage{multirow}
\usepackage{pbox}
\usepackage{xspace}
\usepackage{soul}
\usepackage{tabularx}
\usepackage{amsmath, amsthm, amssymb, amsfonts}
\naaclfinalcopy 
\newcommand{\promote}{\textbf{promote}\xspace}
\newcommand{\suppress}{\textbf{suppress}\xspace}

\newcommand{\good}{\textbf{good}\xspace}
\newcommand{\bad}{\textbf{bad}\xspace}

\newcommand{\support}{\emph{support}\xspace}
\newcommand{\rebuttal}{\emph{rebuttal}\xspace}
\newcommand{\undercut}{\emph{undercut}\xspace}

\newcommand{\other}{\emph{OTHER}\xspace}

\usepackage{enumitem,kantlipsum}
\usepackage{mwe}
\usepackage{subfig}

\title{A Corpus of Deep Argumentative Structures \\as an Explanation to Argumentative Relations}

\author{Paul Reisert$^1$ \and Naoya Inoue$^2$ \and Naoaki Okazaki$^3$ \and Kentaro Inui$^{1,2}$\\
  $^1$ RIKEN Center Advanced Intelligence Project\\
  $^2$ Tohoku University\\
  $^3$ Tokyo Institute of Technology\\
  {\tt paul.reisert@riken.jp, \{naoya-i, inui\}@ecei.tohoku.ac.jp,}\\
  {\tt okazaki@c.titech.ac.jp}
}
\date{}

\begin{document}

\maketitle

\begin{abstract}
In this paper, we compose a new task for deep argumentative structure analysis that goes beyond shallow discourse structure analysis. The idea is that argumentative relations can reasonably be represented with a small set of predefined patterns. For example, using value judgment and bipolar causality, we can explain a support relation between two argumentative segments as follows: Segment 1 states that something is good, and Segment 2 states that it is good because it promotes something good when it happens. We are motivated by the following questions: (i) how do we formulate the task?, (ii) can a reasonable pattern set be created?, and (iii) do the patterns work? To examine the task feasibility, we conduct a three-stage, detailed annotation study using 357 argumentative relations from the argumentative microtext corpus, a small, but highly reliable corpus. We report the coverage of explanations captured by our patterns on a test set composed of 270 relations. Our coverage result of 74.6\% indicates that argumentative relations can reasonably be explained by our small pattern set. Our agreement result of 85.9\% shows that a reasonable inter-annotator agreement can be achieved. To assist with future work in computational argumentation, the annotated corpus is made publicly available.
\end{abstract}

\section{Introduction}
\label{sec:intro}

In recent years, there has been growing interest in the automatic analysis of argumentative texts~\cite{Lippi2015}, such as the identification of a \emph{shallow} discourse structure for such texts by way of argumentative relation detection~\cite[etc.]{Cabrio2012,Stab2014,peldszus2015annotated} and argumentative zoning~\cite[etc.]{teufel2009towards,Levy2014,Houngbo2014}. Argumentative relation detection is the task of identifying argumentative relations (typically, support and attack relations) between discourse segments. Argumentative zoning is the task of identifying argumentative zones such as premise and claim.

Suppose we are analyzing the following argumentative text discussing the topic ``\emph{Should shopping malls be allowed to be open on holidays?}'':
\enumsentence{
  {\it [I as an employee find it practical to be able to shop on weekends.]$_{S_1}$ [Sure, other people have to work on the weekend,]$_{S_2}$ [but they can have days off during the week.]$_{S_3}$} \label{ex:deb1a1}
  }
In this text,\footnote{Slightly modified version of text \texttt{b015} in \cite{peldszus2015annotated}'s argumentative microtext corpus.} segment $S_2$ attacks segment $S_1$ and segment $S_3$ attacks, or undercuts, the relationship between $S_2$ and $S_1$ (argumentative relation detection). In another view, $S_2$ and $S_3$ serve as premises and $S_1$ as a claim (argumentative zoning).

The design of these shallow discourse analysis tasks has an advantage in their simplicity, which makes human annotation simple and reliable, achieving relatively high inter-annotator agreement~\cite{prasad2008penn,Habernal2014,Stab2014,Peldszus2015,rinott2015show,reedacl2016}. Previous studies on discourse analysis, including discourse theories such as Rhetorical Structure Theory~\cite{mann1987rhetorical}, thus have mainly focused on creating corpora for the identification of shallow discourse structures.

In this work, we propose a task design for going beyond shallow discourse structure by analyzing argumentative texts at a deeper level. We consider the task of producing an explanation as to \emph{why} it makes sense to interpret each support/attack relation (i.e. the author's logical reasoning) underlying a given argumentative text. For instance, in Example~\ref{ex:deb1a1}, a reasonable explanation why $S_2$ can be interpreted as an attack to $S_1$ is the following:

\enumsentence{(i) $S_1$ states that \emph{be able to shop on weekends} (relevant to the topic) is a good thing. (ii) $S_2$ presupposes that \emph{be able to shop on weekends} will make \emph{other people work on the weekends}, (iii) which is a bad consequence. (iv) So, $S_1$ states a good aspect of one thing whereas $S_2$ states a bad consequence of the same thing; therefore, $S_2$ attacks $S_1$. \label{ex:2}
}

This direction of task design has several advantages. First, understanding the logical reasoning behind an argumentative text contributes toward determining implicit argumentative relations not indicated with an explicit discourse marker. Analysis of implicit discourse relations is a long-standing open problem in discourse analysis~\cite[etc.]{Ji2015,Chen16}. We expect that this direction of research will provide a new approach to it. Second, identifying the logical reasoning will be useful for a range of argumentation mining applications. One obvious example is to aggregate multiple arguments and produce a logic-based abstractive summary. It will also be required in automatically assessing the quality of the logic structure of a given argumentation (cf. automatic essay scoring~\cite{Song2014,wachsmuthusing}). Furthermore, it will be useful for generation of rebuttals in application contexts where a human and machine are cooperatively engaged in a debate (for decision support or education). Shallow discourse structure analysis, as assumed in previous work, suffers a large gap between what it produces and what is required in these useful potential applications.

In this paper, we address the following research questions:
\begin{enumerate}
\item How do we formulate the task? 
\item Can a reasonable set of patterns can be created?
\item Do the predefined patterns work?
\end{enumerate}

Our contributions are as follows: (i) we compose a list of predefined patterns for explaining argumentative relations that cover a majority of explanations; (ii) we conduct a detailed, three-step annotation process which enables us to keep track of the annotation results; and (iii) we create the first, publicly available corpus to incorporate a deep structure annotation for argumentative relations that achieves good annotator agreement.

\section{Related Work}
\label{sec:related_work}

Conventionally, discourse structure analysis has been studied in the context of discourse relation identification. Earlier work includes discourse theories such as rhetorical structure theory which aim at creating coherent, tree-like structures for describing texts, where text units are typically adjacent~\cite{mann1987rhetorical}. Other theories such as cross-document structure theory focus on the identification of discourse relations spanned across multiple documents~\cite{radev2000common}. The Penn Discourse TreeBank, the largest manually annotated corpus for discourse relations, targeted both implicit and explicit relation detection for either adjacent sentences or clauses~\cite{prasad2008penn}. In the field of Argumentation mining, previous work has proposed several kinds of tasks such as structure identification task (e.g., support-attack relation detection)~\cite{Cabrio2012,Peldszus2015}. In addition, a wide variety of corpora have been created in several domains including scientific articles, essays, and online discussions~\cite{Habernal2014,Stab2014}. These studies aim to capture the shallow structures of arguments and do not try to explain a writer's reasoning.

One may think that stance classification is closely related to our task in terms that identifying the stance of a debate participant towards a discussion topic at a document or sentence level, and several corpora have been created for the task~\cite[etc.]{Hasan2014,Persing2016}. However, since this direction of research focuses only on the classification of the stance polarity of a given paragraph or sentence, generating the explanation between two argumentative components has been out of scope. 

Several argumentative corpora have been created for argumentation mining fields such as argument component identification, argument component classification, and structure identification~\cite{Reed2008,rinott2015show,Stab2014}, but none of them are like our current task setting. \shortcite{Reed2008} annotated AraucariaDB corpus~\cite{Reed2005} with Walton's Argumentation Scheme, and the successive work~\cite{Feng2011} creates a machine learning-model to classify an argument into five sets of schemes. However, they do not annotate instantiations of variables in Argumentation Scheme, and do not report the inter-annotator agreement.
\cite{green2015identifying} conducted preliminary work on identifying a set of argumentation schemes used in scientific articles based on Argumentation Scheme. However, they do not actually create a corpus. 

\section{Explaining argumentative relations}
\label{sec:patterns}

The key challenge of defining a computationally feasible explanation generation task is to establish the machine-friendly representation of \emph{explanation of argumentative relation} (EAR). In Argumentation Theory, a number of formal theories to describe an argumentative structure of a debate have been studied~\cite[etc.]{hastings1963reformulation,perelman1971new,walton2008argumentation}. However, it is not trivial how to operationalize such theories as a computational task. The main focus of the theories is purely in organizing the ``nature'' of human argumentation, where the level of machine-friendliness is not necessary.

\begin{figure*}[t]
      \centering
      \includegraphics[clip,bb=0 80 960 540, scale=0.5]{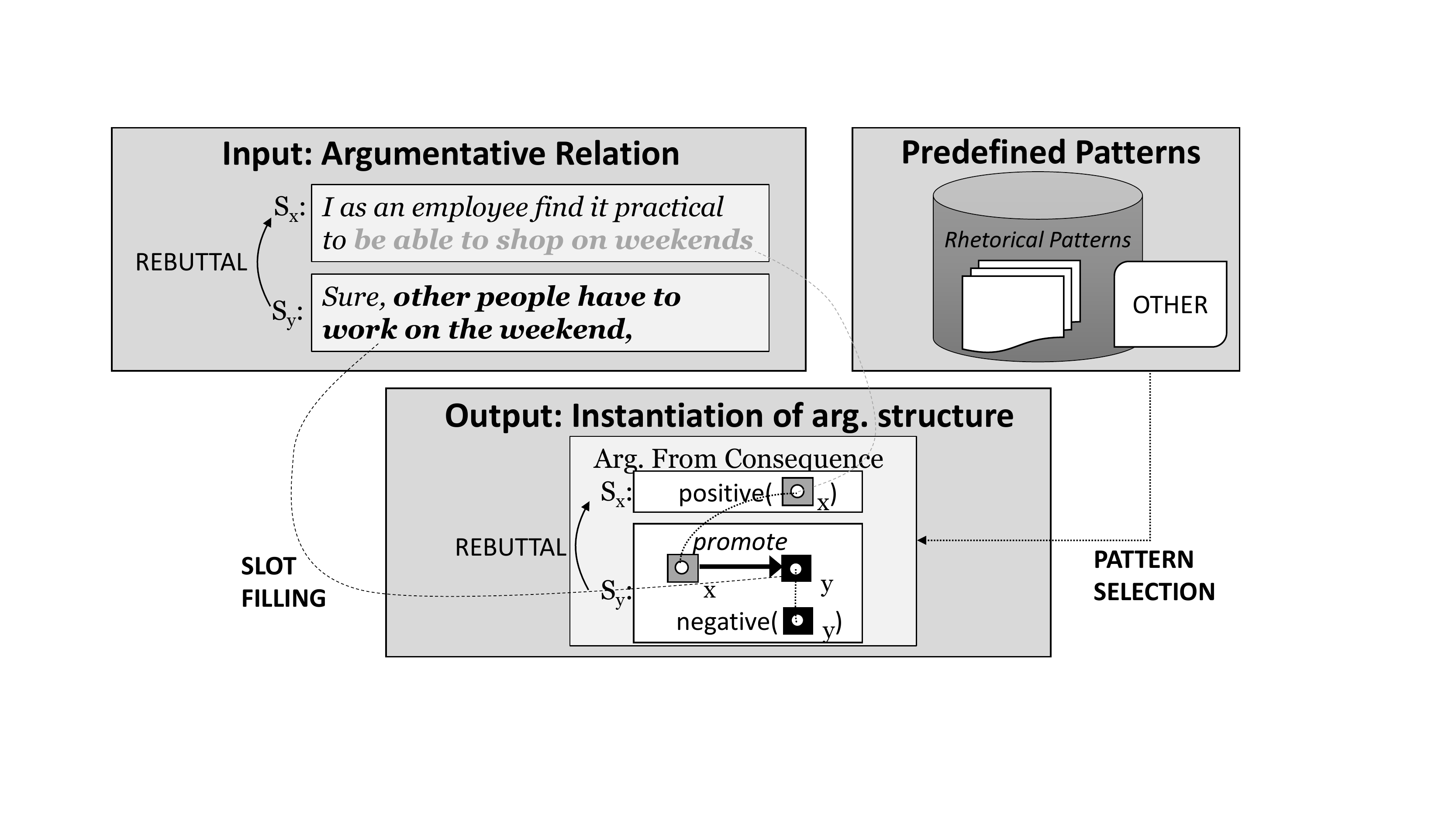}
      \caption{Overview of our framework proposed for the task of deep argumentative relation analysis.}
      \label{fig:overview}
\end{figure*}

To address this issue, we simplify the EAR generation task as the task of identifying a pattern of explanation from a predefined pattern set (henceforth, \emph{rhetorical pattern}) coupled with a slot-filling problem, where the slot is linked with a phrase in an input text. Intuitively, to generate an explanation to the rebuttal relation from $S_2$ to $S_1$ in Example \ref{ex:deb1a1}, the first step is to identify a rhetorical pattern ``\emph{$S_1$ states that $x$ is good, but $S_2$ states that $x$ is bad because $y$ is bad, and when $x$ happens, $y$ will be promoted.}'', as illustrated in Figure~\ref{fig:overview} (see $\rebuttal(S_1,S_2) \leftarrow ...$). The second task is to fill the slots $x,y$ with a phrase from the text: $x=$``\emph{be able to shop on weekends}'' and $y=$``\emph{other people have to work on the weekend}''.

Our hypothesis is that such rhetorical patterns are not arbitrary but highly skewed. We thus create an inventory of major rhetorical patterns and annotate only typical patterns of logical reasoning with them, leaving uncommon patterns to be labeled simply as ``OTHER'' . In fact, as we report in Section~\ref{sec:corpus_study}, the variety of logical reasoning underlying argumentative relations can be largely captured by only the small number of predefined patterns in the corpus we use. These design decisions make our task of identifying deep argumentative structure simple while going beyond previous task settings of shallow analysis.

\noindent\textbf{Argumentative relations} We adopt the definitions of \support, \rebuttal, and \undercut relations from \cite{peldszus2015annotated}. The relations are defined as follows:
\begin{itemize}[labelwidth=!, labelindent=5pt]
  \item \support: One argumentative segment supports the acceptability of another argumentative component.
  \item \rebuttal: One relation refutes (attacks) the acceptability of another argumentative component.
  \item \undercut: One argumentative component attacks the link (relation) between a given argumentative relation.
\end{itemize}

\noindent\textbf{{Rhetorical patterns}} We consider building our inventory of rhetorical patterns based on \cite{walton2008argumentation}'s Argumentation Schemes, one prominent argumentation theory. Among their 65 schemes, we create our first rhetorical patterns from the \emph{Argument from Consequences} (AC) scheme: 
\begin{itemize}[labelwidth=!, labelindent=5pt]
  \item Premise: If $x$ is (not) brought about, \textit{good} (\textit{bad}) consequences $y$ will occur.
  \item Conclusion: $x$ should (not) be brought about.
\end{itemize}

We analyze the argumentative microtext corpus~\cite{peldszus2015annotated}, a small, highly reliable corpus consisting of important ingredients for computational argumentation, and find that the AC scheme and its variants are commonly used in argumentation (see Section~\ref{sec:setup} for details). The rhetorical patterns created for \rebuttal and \undercut are closely related to the Critical Questions in Argumentation Schemes, which assess the quality of argumentation (e.g. \emph{Are there other consequences of the opposite value that should be taken into account?}).

First, we create the rhetorical pattern of \support from $S_2$ to $S_1$ as $S01$ in Figure~\ref{fig:patterns1}. $S01$ is interpreted as:
%
\begin{itemize}[labelwidth=!, labelindent=5pt]
\item \textbf{S01}: $S_1$ states that $x$ is \textbf{good}, and $S_2$ states that $x$ is \textbf{good} because $y$ is \textbf{good} and when $x$ happens (or happened), $y$ will be (or was) \textbf{promoted} (or \textbf{not suppressed}).
\end{itemize}
We then extend this pattern, altering its value judgment (i.e. \good or \bad) and causality direction (i.e. \promote or \suppress) for the remaining relations.\footnote{The causality and value judgment may both either be explicitly stated in the text or implicitly stated.} For example, the following interpretation explains $S02$:
\begin{itemize}[labelwidth=!, labelindent=5pt]
\item \textbf{S02}: $S_1$ states that $x$ is \textbf{good}, and $S_2$ states that $x$ is \textbf{good} because $y$ is \textbf{good} and when $x$ does not happen (or didn't happen)\footnote{$x$ not happening or did not happen is signified by $\neg{x}$.}, $y$ will be (or was) \textbf{promoted} (or \textbf{not suppressed}).
\end{itemize}

For the \rebuttal relation, represented in patterns $R01$ through $R08$, the value judgment and causality direction in the patterns are modified to fit each relation. Thus, $R01$ can be interpreted as:
\begin{itemize}[labelwidth=!, labelindent=5pt]
\item \textbf{R01}: $S_1$ states that $x$ is \textbf{good}, but $S_2$ states that $x$ is \textbf{bad} because $y$ is \textbf{good} and when $x$ happens (or happened), $y$ will be (or was) \textbf{suppressed} (or \textbf{not promoted}).
\end{itemize}
In Figure~\ref{fig:overview}, $\emph{rebuttal}(S_1, S_2)$ is explained by pattern $R03$, with $x$=``\textit{be able to shop on weekends}'', something \good, and $y$=``\textit{other people work on the weekend}'', something \bad. 

For the \undercut relation, represented in patterns $U01$ through $U08$, the value judgment and causality direction in the patterns are also modified to fit each relation. The following example shows the interpretation of the \undercut relations in Figure~\ref{fig:patterns1} (e.g. $U01$):

\begin{itemize}[labelwidth=!, labelindent=5pt]
\item \textbf{U01}: $R_1$ supports the goodness of $x$, but $S_1$ states that $x$ is \textbf{bad} because $y$ is \textbf{good} and when $x$ happens (or happened), $y$ will be (or was) \textbf{suppressed} (or \textbf{not promoted})
\end{itemize}
, where $R_1$ is a given argumentative relation.

\begin{figure*}[t]
      \centering
      \includegraphics[clip, bb=0 200 3973 1430, scale=0.2]{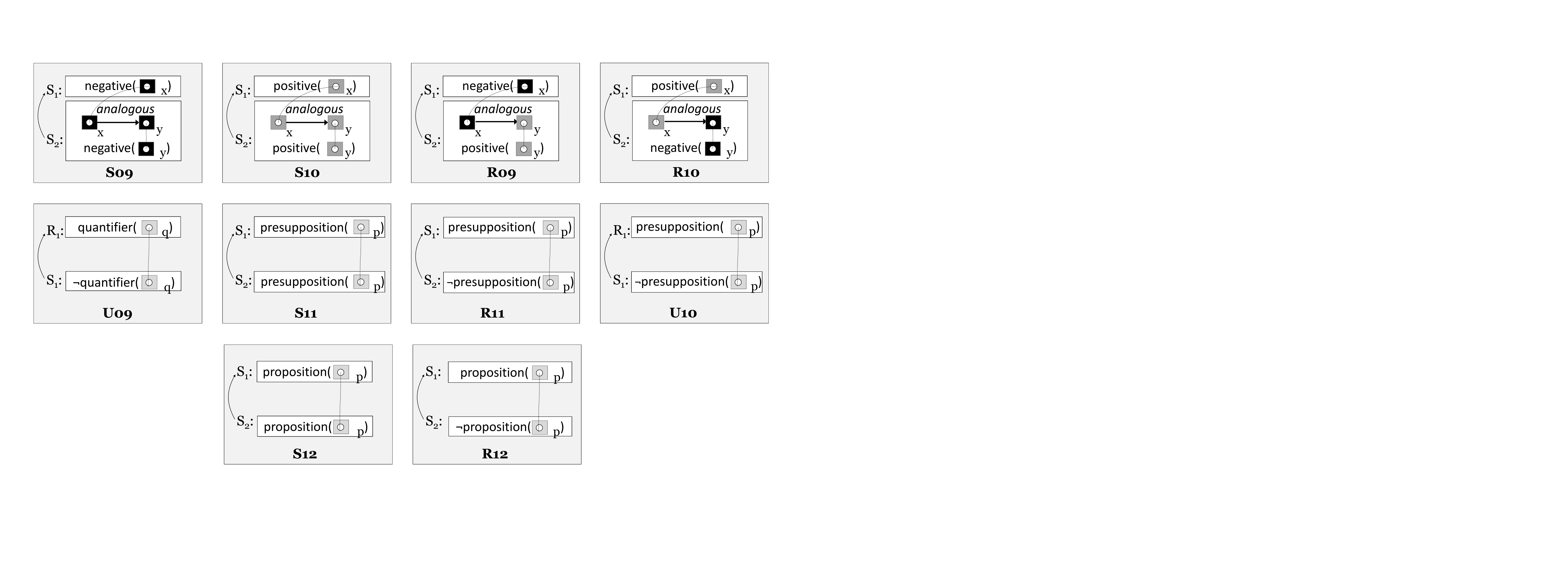}
      \caption{Rhetorical patterns for argument from analogy ($S09$-$S10$, $R09$-$R10$), quantifier ($U09$), presupposition ($S11$,$R11$,$U10$), and proposition ($S12$,$R12$).}
      \label{fig:patterns2}
\end{figure*}

Additionally, we create a few rhetorical patterns to capture minor non-AC arguments for \support, \rebuttal, and \undercut relations. Figure~\ref{fig:patterns2} shows the patterns for \emph{analogous} and \emph{propositional} explanations for \support and \rebuttal relations ($S09$-$S10$ and $R09$-$R10$, respectively). The pattern is interpreted as follows (e.g. $R09$):
\begin{itemize}
\item \textbf{R09}: $S_1$ states that $x$ is \textbf{bad}, but $S_2$ states that $x$ is \textbf{good} because $y$ is \textbf{good} and is analogous to $x$. 
\end{itemize}
\enumsentence{
  {\it [Since however skat, chess etc. are not accepted as Olympic events,]$_{S_1}$ [computer games should also not be recognized as Olympic events.]$_{S_2}$} \label{ex:analogy}
  }

In Example~\ref{ex:analogy}, \textit{computer games} in $S_2$ is analogously compared to \textit{skat, chess, etc.}, a \bad thing; thus, $S09$ would be selected.

For the \undercut relation, our analysis revealed that a quantifier in a relation could be attacked. Thus, we create the relation $U09$ for \undercut, represented as: 
\begin{itemize}[labelwidth=!, labelindent=5pt]
\item \textbf{U09}: $R_1$ assumes a quantifier $q$, but $S_2$ disagrees with it.
\end{itemize}

\enumsentence{
  {\it [[Intelligent services must urgently be regulated more tightly by parliament;]$_{S_2}$ [this should be clear to everyone after the disclosures of Edward Snowden.]$_{S_3}$]$_{R_1}$ [Granted, those concern primarily the British and American intelligence services,]$_{S_1}$} \label{ex:quantifier}
  }

In Example~\ref{ex:quantifier}, $R_1$, a \support{($S_2$,$S_3$)} relation, assumes that all intelligent services should be regulated more tightly; however, $S_1$ states that only two services are concerned.

To capture the argument where the underlying assumptions in one segment are supported or attacked by another, we introduce the relations $S11$, $R11$, and $U10$ for \support, \rebuttal, and \undercut, respectively. The pattern can be interpreted as follows (e.g. $S11$):
\begin{itemize}[labelwidth=!, labelindent=5pt]
\item \textbf{S11}: $S_1$ assumes a presupposition $p$, and $S_2$ agrees with it.\footnote{We mark only the term presupposition; however, we will write the content as a note of the pattern annotation.}
\end{itemize}

\enumsentence{
  {\it [For dog dirt left on the pavement dog owners should by all means pay a bit more.]$_{S_1}$ [Indeed, it's not the fault of the animals]$_{S_2}$} \label{ex:presupposition}
  }

In Example~\ref{ex:presupposition}, $S_1$ presupposes that it is the fault of the animals, but $S_2$ disagrees. Thus, pattern $R11$ would be selected.\footnote{$\neg$presupposition means that $S_2$ disagrees with the presupposition in $S_1$ ($R_1$ in the case of \undercut). This notion is similar for \emph{quantifier} and \emph{proposition}.}

We also create patterns for propositional explanations, represented in patterns $S12$ and $R12$. The patterns can be interpreted as follows (e.g. $S12$):
\begin{itemize}[labelwidth=!, labelindent=5pt]
\item \textbf{S12}: $S_1$ states a proposition $p$, and $S_2$ restates it.
\end{itemize}

If an argumentative relation cannot be represented by one of the rhetorical patterns above, the relation is labeled as \other.

In total, we create 5 types of rhetorical patterns: \textit{Argument from Consequence}, \textit{Presupposition}, \textit{Argument from Analogy}, \textit{Proposition}, and \textit{Quantifier}.

\section{Corpus study}
\label{sec:corpus_study}

To examine the feasibility of our task, we conduct the corpus study for annotating argumentative relations using our rhetorical patterns. We report the annotation process and a corpus study.
\subsection{Methodology}
\label{sec:setup}

We annotated 89 texts\footnote{The original corpus has 112 texts, but we ignored 23 of the texts which did not include a debate topic question.} from \cite{peldszus2015annotated}'s argumentative microtext corpus\footnote{https://github.com/peldszus/arg-microtexts}, each consisting of roughly five segments composed of a topic question, a main claim and support and attack segments (see Figure~\ref{fig:overview} for a partial example taken from the corpus). 357 argumentative relations between segments have been manually annotated as either \emph{support}, \emph{rebuttal}, or \emph{undercut}, where each relation makes up 62.7\% (224/357), 23.5\% (84/357) and 13.8\% (49/357), respectively.

To study the coverage of rhetorical patterns, we asked annotators to mark a relation as the special pattern \other when the relation cannot be explained by the rhetorical patterns. We divided the corpus into two disjoint sets: (i) a development set (or dev set; 20 texts, 87 relations) and (ii) test set (69 texts, 270 relations). We used only the dev set to induce the rhetorical patterns described in Section~\ref{sec:patterns} and conducted several trial annotations. For an inter-annotator agreement study, we asked two fluent-English speakers to explain each argumentative relation with the rhetorical patterns. On the dev set, we encouraged two annotators to discuss the annotation disagreements with each other.

We use brat~\cite{stenetorp2012brat}, the general-purpose annotation tool for NLP. Given the original, segmented argumentative text, the annotated argumentative relations, and the predefined list of rhetorical patterns, the annotators select an appropriate pattern and mark arbitrary phrases in the segmented text when filling in the pattern slots.

\subsection{Results and discussion}
\label{sec:results}

Although we simplified the task of EAR generation in Section~\ref{sec:patterns}, human annotation may still be inconsistent. We thus adopted a three-step annotation process to maintain the quality of annotation. The final corpus includes the annotations from all these steps and the dev set.
\begin{figure*}[t]
\centering
\begin{minipage}{.5\linewidth}
\centering
\subfloat[]{\label{main:a}\includegraphics[bb=0 0 541 220, scale=.55]{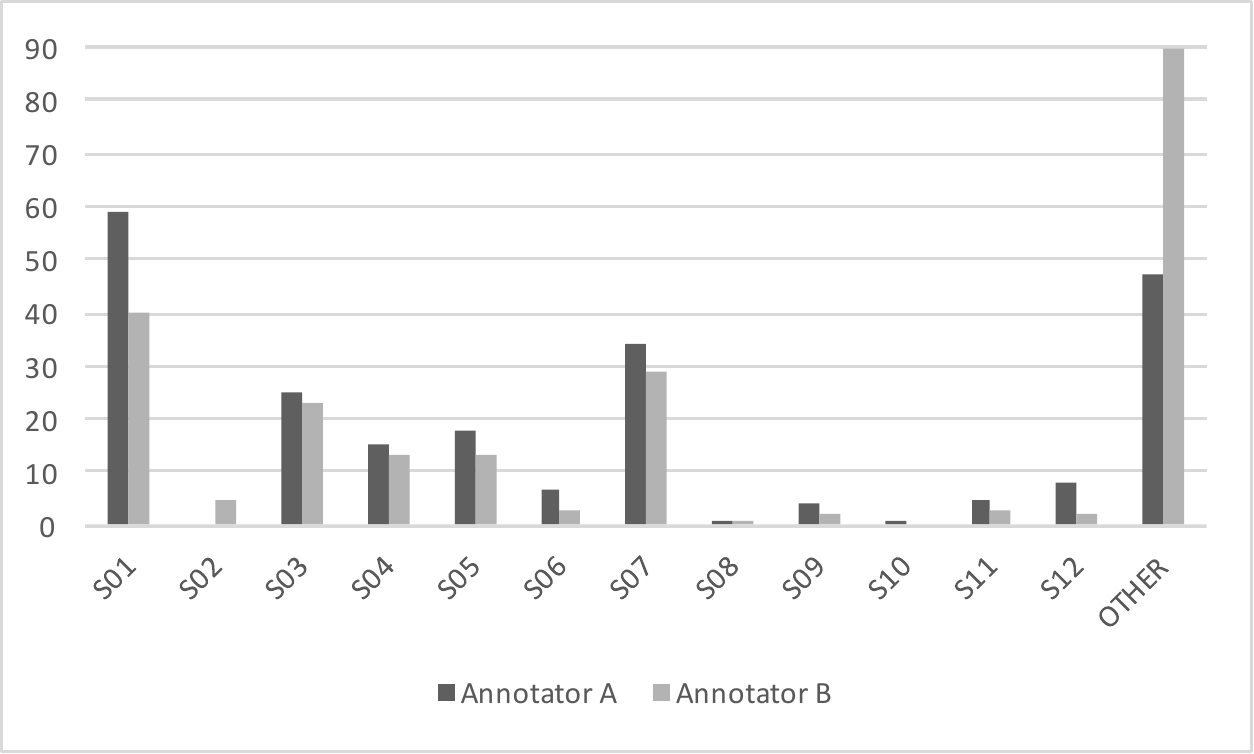}}
\end{minipage}%
\begin{minipage}{.5\linewidth}
\centering
\subfloat[]{\label{main:b}\includegraphics[bb=0 0 541 220, scale=.55]{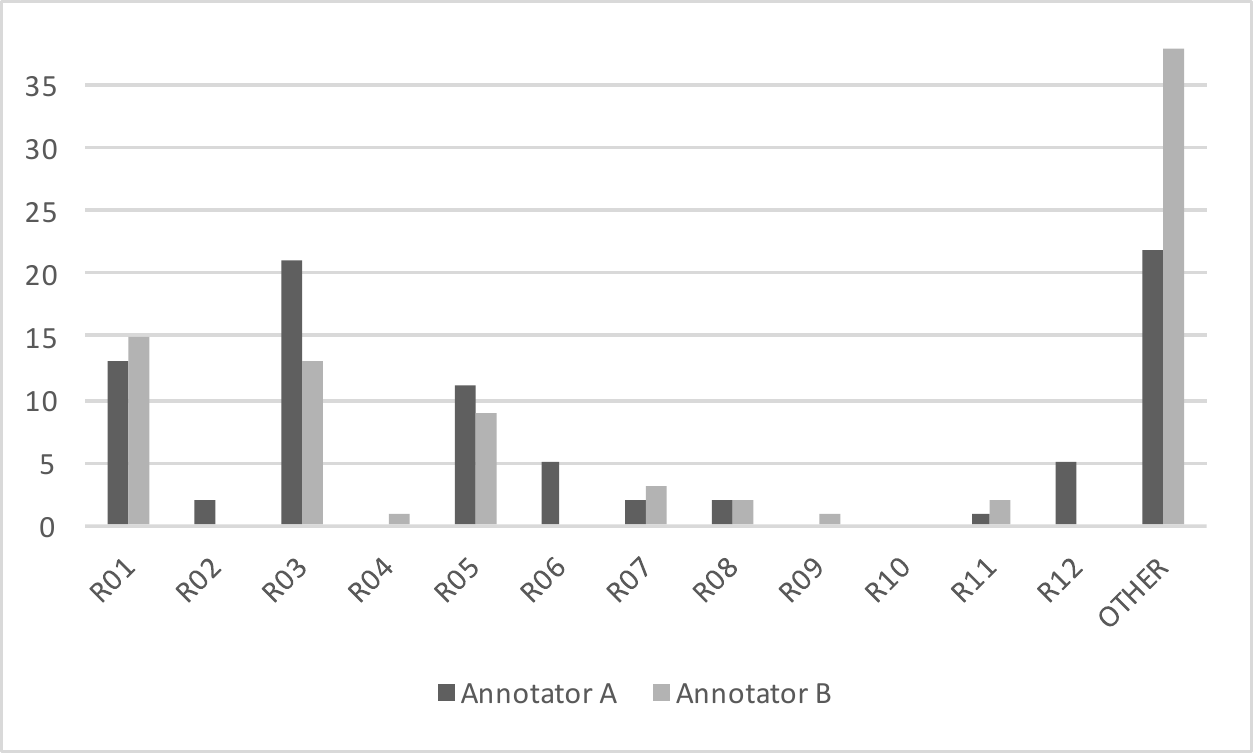}}
\end{minipage}\par\medskip
\centering
\subfloat[]{\label{main:c}\includegraphics[bb=0 0 541 220, scale=.55]{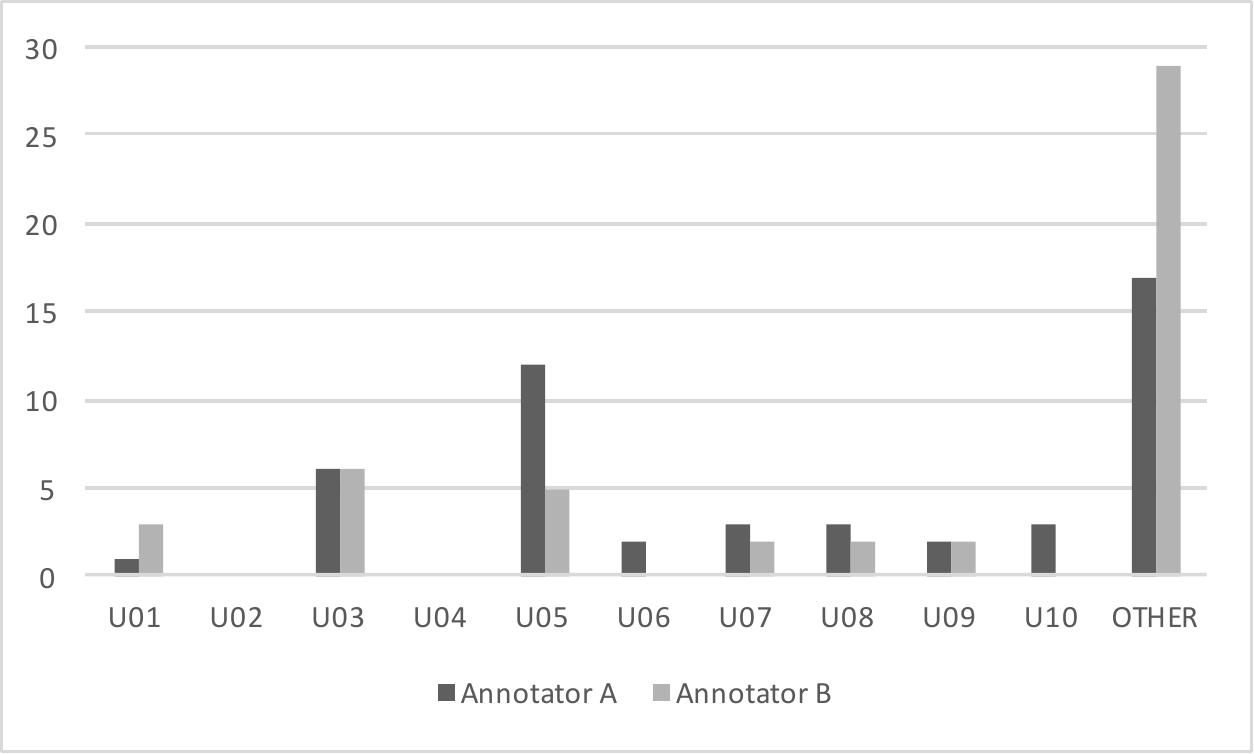}}

\caption{Distribution of agreements for support (a), rebuttal (b), and undercut (c) relations between two annotators at Stage 1.}
\label{fig:basic}

\end{figure*}

\noindent\textbf{Stage 1.} First, we had the annotators individually annotate the test set. The distribution of annotations between annotators for all relations can be seen in Figure~\ref{fig:basic}. The results appear consistent; however, one annotator was more biased towards labeling \other and the first pattern for each relation. To analyze the inter-annotator agreement, we categorized each pair of EAR annotations as ``\emph{agreed}'' if (i) the rhetorical patterns selected by both are exactly the same or semantically similar (e.g. pattern (1) and (2) in Section~\ref{sec:patterns}) \emph{and} (ii) the phrases associated with the pattern slots are exactly the same or overlapped. For judging two patterns as semantically similar, we consider whether the value judgment of both $x$ and $y$ for the patterns are equivalent. For example, from our pattern list in Section~\ref{sec:patterns}, $S01$ and $S02$ would be semantically equivalent.

In total, the observed agreement for relations categorized as ``agreed'' was 46.3\% (125/270). Simultaneously, the Cohen's Kappa score for the selected rhetorical patterns, which we calculate using selected rhetorical patterns only, where we consider semantically similar patterns as equivalent, was 0.45 which indicates a moderate agreement.\footnote{The Cohen's Kappa score is calculated using relation-only agreement for all stages of our corpus study.} For ``agreed'' instances, we found that the coverage of rhetorical patterns (i.e. the number of EARs representable by at least one of our patterns) is 68.0\% (85/125), which supports our ``majority pattern'' hypothesis on this corpus.

\begin{figure*}[t]
      \centering
      \includegraphics[clip, bb=0 0 960 350, scale=0.6]{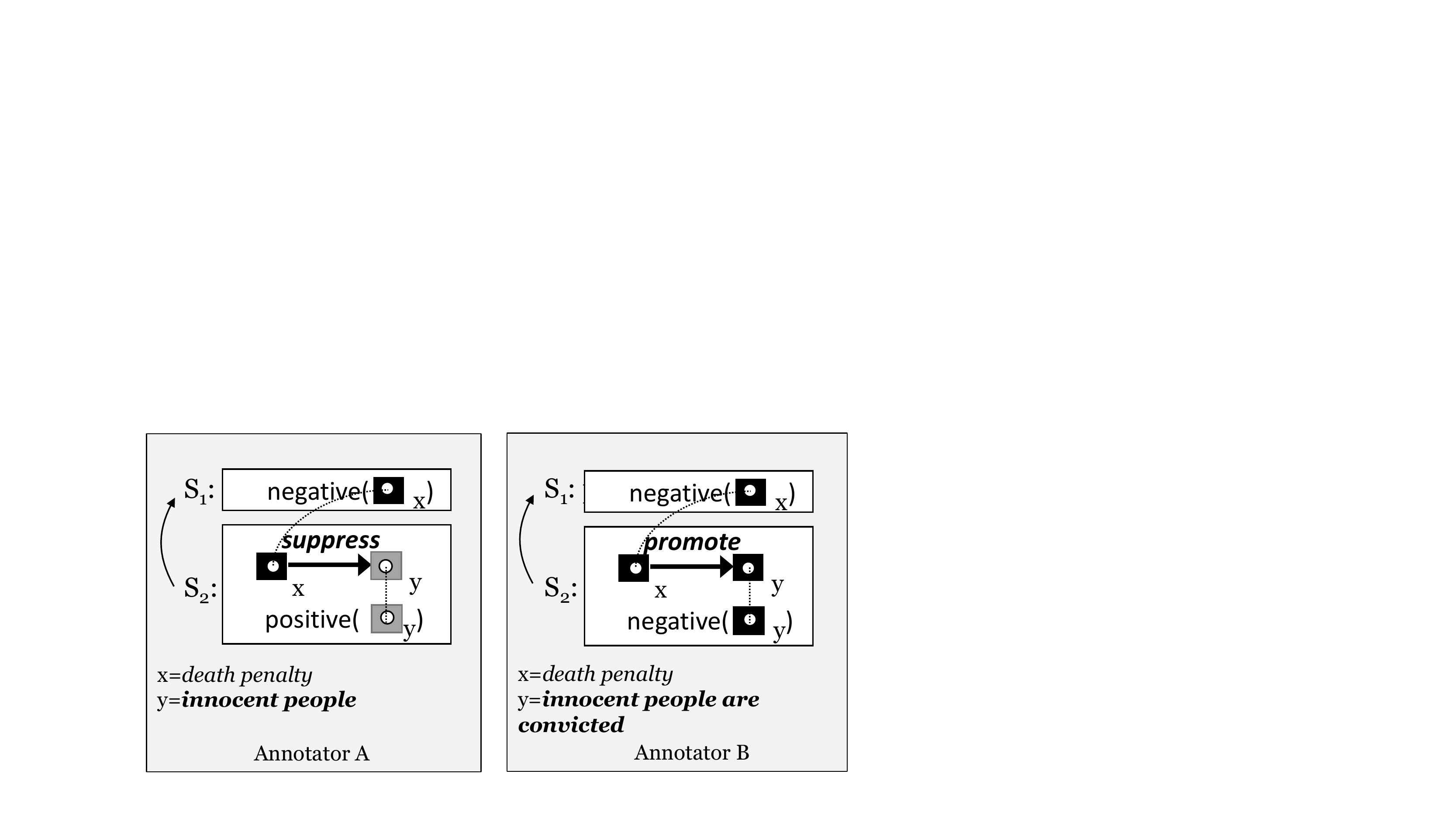}
      \caption{Multiple ways to interpret a \support{($S_1$,$S_2$) relation.}}
      \label{fig:error}
\end{figure*}

\noindent\textbf{Stage 2.}
Due to the complexity of the EAR generation task, there is a possibility that agreeable annotations among the 145 disagreements from Stage 1 exist. Therefore, for each disagreement, we individually asked annotators whether they agree with the other's annotation or are unsure about it.
The cross-check revealed that 54.0\% (78/145) of the disagreements were agreeable and mainly caused by multiple ways to interpret a relation or small mistakes in an annotation (e.g. filling a slot with a phrase from an adjacent segment's text) made by one annotator, which could be reasonably considered as ``\emph{semi-agreed}''. Figure~\ref{fig:error} shows an example of the relation \support{($S_1$,$S_2$)} between the following two segments, where two different interpretations were agreed by each other at Stage 2.
\begin{itemize}
\item $S_1$: \textit{The death penalty should be abandoned everywhere.}
\item $S_2:$ \textit{Sometimes innocent people are convicted.}
\end{itemize}

The figure indicates that both annotators agreed that the \textit{Argument from Consequences} pattern was appropriate, and $x$ is \emph{death penalty}. However, one annotator labeled $y$ as \textit{innocent people}, a \good thing, and the other annotator labeled $y$ as \textit{innocent people are convicted}, a \bad thing, which was considered a disagreement at Stage 1. The annotators agreed with each other's annotation because the suppression of \emph{innocent people} and the promotion of \emph{innocent people are convicted} are semantically compatible. This kind of annotation disagreement is considered as ``semi-agreed'' at Stage 2.

In total, 75.2\% (203/270) of the relations are categorized as ``semi-agreed'', indicating a higher coverage of 75.9\% (154/203), which further supports our hypothesis. Our Cohen's Kappa score at this stage is 0.71 which indicates a good agreement. Note that the annotators are not biased toward saying ``agree with the other's annotation'': one annotator said 56 yes, 69 no and 21 unsure, and the other said 55 yes, 52 no and 39 unsure.

\begin{figure*}[t]
\centering
\begin{minipage}{.5\linewidth}
\centering
\subfloat[]{\label{main:a}\includegraphics[bb=0 0 360 217, scale=.55]{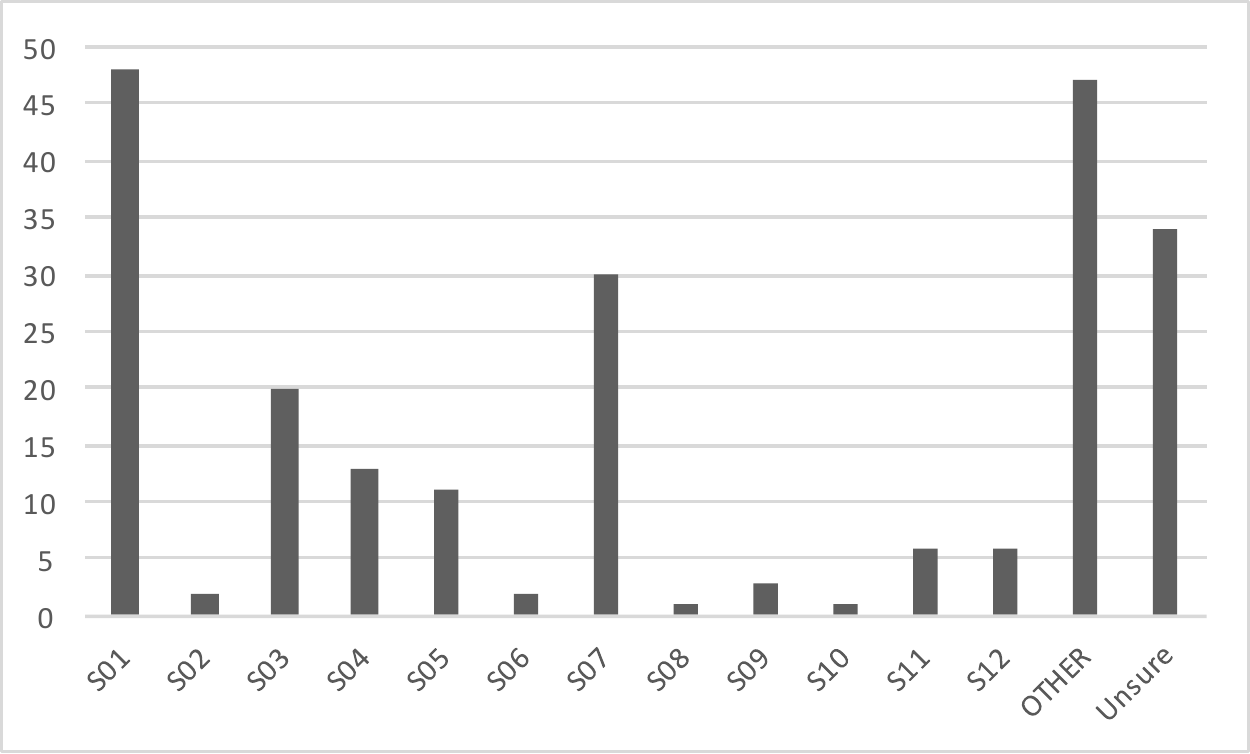}}
\end{minipage}%
\begin{minipage}{.5\linewidth}
\centering
\subfloat[]{\label{main:b}\includegraphics[bb=0 0 360 217, scale=.55]{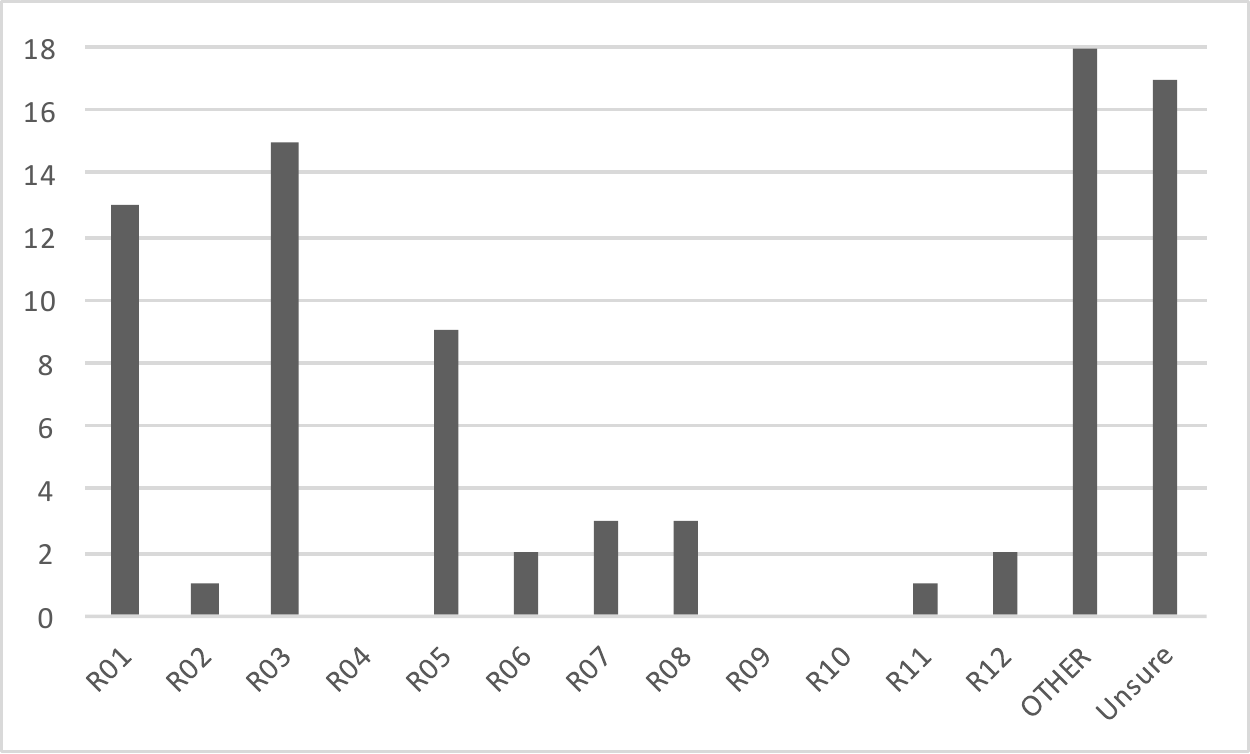}}
\end{minipage}\par\medskip
\centering
\subfloat[]{\label{main:c}\includegraphics[bb=0 0 360 217, scale=.55]{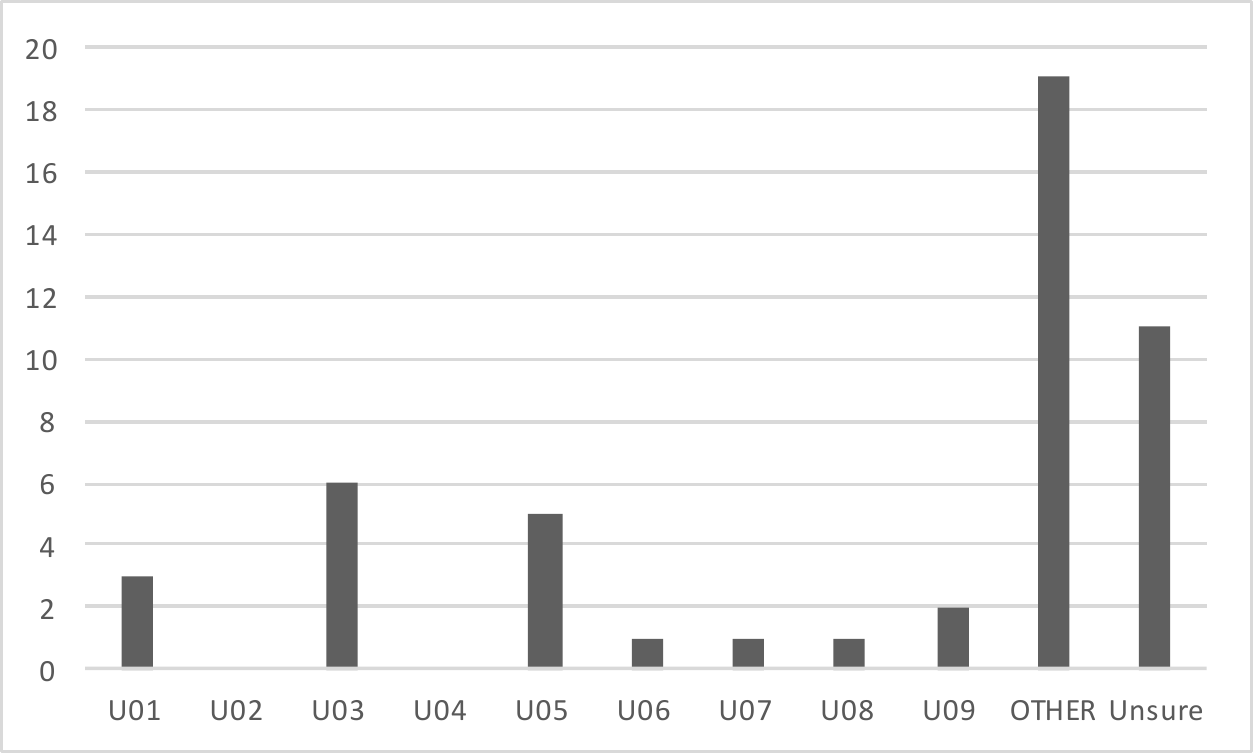}}

\caption{Distribution of final corpus statistics for support (a), rebuttal (b), and undercut (c) relations.}
\label{fig:main}

\end{figure*}

\noindent\textbf{Stage 3.}
We further asked both annotators to discuss the remaining 67 disagreements (i.e. both said no) or unsure instances (out of 146) from Stage 2 with each other to see if an agreement (or disagreement, for unsure instances) could be reached. After the discussion, 85.9\% (232/270) of the relations are categorized as ``\emph{semi-agreed}''. The Cohen's Kappa score at this stage is 0.80, indicating a good agreement. The coverage of patterns is 74.6\% (173/232), which is slightly lower than that of Stage 2 due to less ambiguity. Given that an argumentative relation could have two agreeable annotations, we randomly choose an agreement per relation. Figure~\ref{fig:main} shows the distribution of agreements for \support, \rebuttal, and \undercut relations in our final corpus.

The disagreements that turned out to be agreeable (i.e. one of annotators changed his mind from no to yes) were mainly caused by an underspecification in the guidelines. For example, one annotator sometimes blindly believed that a rhetorical pattern should be annotated when it is consistent with a commonsense knowledge, regardless of whether it actually represents the writer's logic. Example~\ref{ex:logic_debate} shows an instance where one annotator selected an argument from consequence rhetorical pattern, where $y$=\textit{"a murderer"} and is suppressed by $x$=\textit{"death penalty"}, because they believed that their annotation is common knowledge.

\enumsentence{
  {\it [The death penalty is a legal means that as such is not practicable in Germany.]$_{S_1}$[Even if many people think that a murderer has already decided on the life or death of another person,]$_{S_2}$
  }\label{ex:logic_debate}
}

For the remaining 38 instances (out of 67), we found that some argumentative texts were not clear enough to understand. We speculate that the precise meaning of the original text was lost in some cases because the corpus is translated from German, preserving its argumentative structure (although it is professionally translated).

\section{Further Analysis}
Towards creating a computational model to identify deep argumentative structure, we conducted a subjective, preliminary analysis consisting of manually analyzing contextual clues or world knowledge which is required to identify rhetorical pattern and its slot instances. For this analysis, we examined 50 random relations from our corpus agreed upon in Stage 3 in Section~\ref{sec:corpus_study}.\footnote{We ignore agreements labeled as \other for this analysis.}

\begin{table*}[t]
\center
\caption{Distribution of clues required for capturing deep argumentative structure.}
\label{tbl:clues}
\begin{tabular}{l|r|r}
\hline\hline
Pattern & Instances & Clues\\\hline
Argument from Consequence & 44 & Value Judgment (35/44), Causality (13/44)\\
Argument from Analogy & 1 & - \\
Presupposition & 1 & - \\
Proposition & 2 & Value Judgment (2/2) \\
Quantifier & 2 & - \\\hline
\end{tabular}
\end{table*}

Table~\ref{tbl:clues} shows the distribution of pattern types and clues from our analysis. We observed several contextual keywords important for both value judgment and bipolar causality. In Example~\ref{exx}, we found the phrases \textbf{should not be left to} and \textbf{escalation} to be important for the value judgment of $S_1$ and $S_2$, respectively, and we found the phrase \textit{for that course can lead to} important for causality.
\enumsentence{
  [The developments in that conflict \textbf{should not be left to} former Cold War opponents alone,]$_{S_1}$[\textit{for that course can only lead to} \textbf{escalation} in some form.]$_{S_2}$
  \label{exx}
  }
However, we observed most patterns required world knowledge implicitly assumed by the writers. For the topic \textit{Should Germany introduce the death penalty?}, Example~\ref{exxx} shows that the causality between the segments $S_1$ and $S_2$ in the relation \support{($S_1$,$S_2$)} is implicit, where the slots are written in bold.
\enumsentence{
  [One should not re-introduce \textbf{capital punishment} in Germany]$_{S_1}$[Everyone must be given \textbf{the chance to hone their conscience and possibly make amends for their deed}]$_{S_2}$\label{exxx}
  }
Likewise, in Example~\ref{exxxx}, the phrase \textit{Owner-run shops}, which has no contextual clues for its value judgment, is annotated as \good by our annotators because it is commonsense knowledge that such shops are typically a \good thing.
\enumsentence{
  [Owner-run shops may potentially be overwhelmed by additional work times on Sundays and holidays]$_{S_1}$\label{exxxx}
  }

\section{Conclusion and Future Work}
\label{sec:conclusion}
We composed a new task for deep argumentative structure analysis. We developed a small list of predefined patterns for explaining argumentative relations (rhetorical patterns) and conducted the corpus study. Although we created our corpus on top of an existing argumentation corpus with several texts, our results indicate that the rhetorical patterns are highly skewed, and even with a small set of predefined patterns, we can cover a majority of explanations (up to 74.6\%). We believe that the design decision to leave a wide variety of long-tailed minor classes of explanations as \other helps keep the generation task simple. Furthermore, our results can be considered a good achievement by discussing disagreements (up to 85.9\%), considering that this is a generation task, which generally indicates a bad inter-annotator agreement. The annotated corpus is made publicly available\footnote{https://github.com/preisert/deep-arg-structure-corpus}, which should facilitate the research of deep argumentation structure analysis. In our future work, we plan to test our coverage in different domains.

\textbf{Acknowledgements}
This work was supported by JSPS KAKENHI Grant Numbers 15H01702, 16H06614, and JST CREST Grant Number JPMJCR1513.

\appendix

\begin{figure*}[t]
      \centering
      \includegraphics[clip, bb=0 0 1992 2281, scale=0.2]{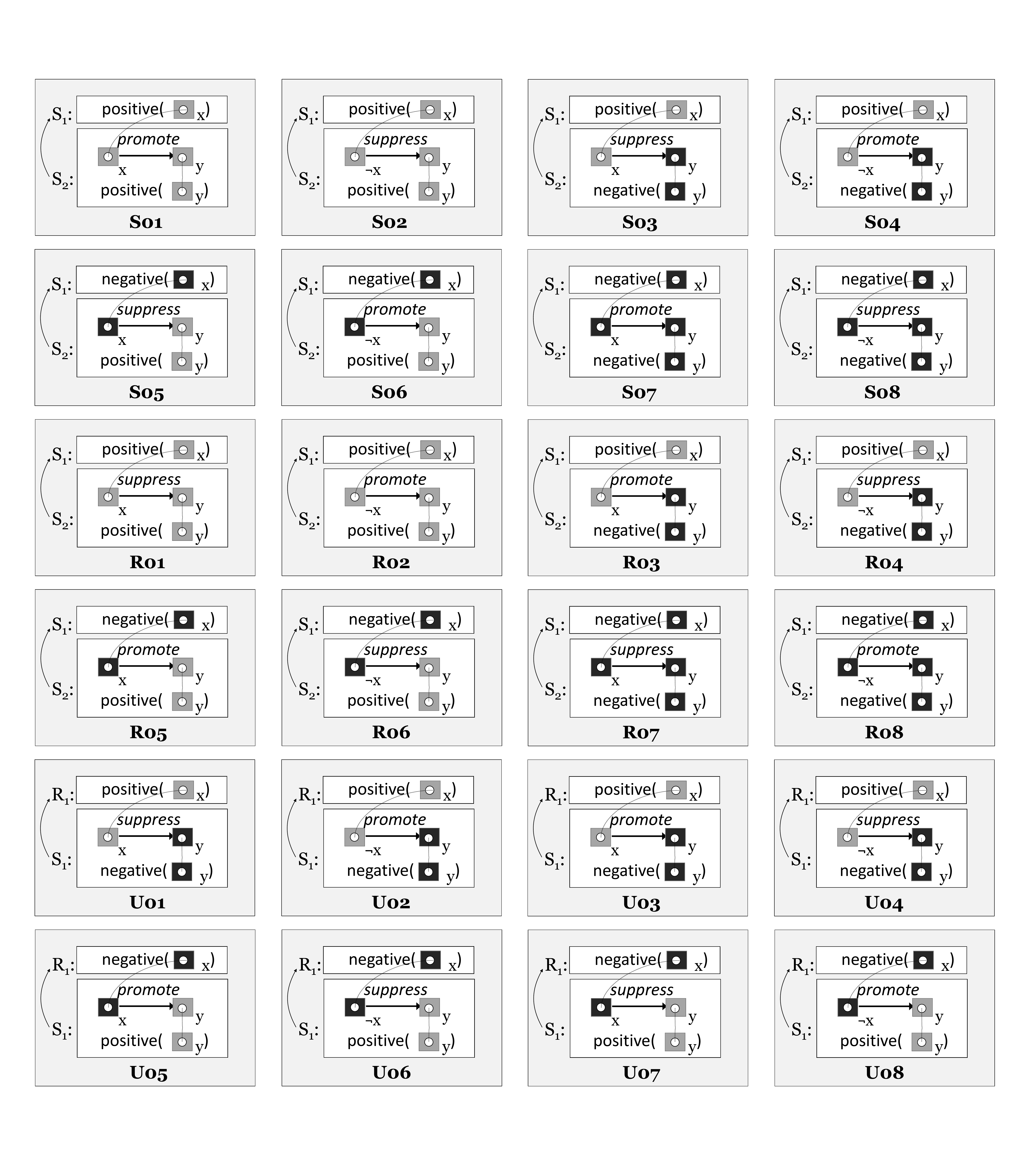}
      \caption{Rhetorical patterns representing the argument from consequences scheme.}
      \label{fig:patterns1}
\end{figure*}

\bibliographystyle{naaclhlt2016}
\bibliography{naaclhlt2016}

\begin{thebibliography}{}

\bibitem[\protect\citename{Cabrio and Villata}2012]{Cabrio2012}
Elena Cabrio and Serena Villata.
\newblock 2012.
\newblock {Generating abstract arguments: A natural language approach}.
\newblock In {\em Frontiers in Artificial Intelligence and Applications},
  number~1, pages 454--461.

\bibitem[\protect\citename{Chen \bgroup et al.\egroup }2016]{Chen16}
Jifan Chen, Qi~Zhang, Pengfei Liu, Xipeng Qiu, and Xuanjing Huang.
\newblock 2016.
\newblock Implicit discourse relation detection via a deep architecture with
  gated relevance network.
\newblock In {\em Proceedings of the 54th Annual Meeting of the Association for
  Computational Linguistics}, pages 1726--1735. Association for Computational
  Linguistics.

\bibitem[\protect\citename{Feng and Hirst}2011]{Feng2011}
Vanessa~Wei Feng and Graeme Hirst.
\newblock 2011.
\newblock Classifying arguments by scheme.
\newblock In {\em Proceedings of the 49th Annual Meeting of the Association for
  Computational Linguistics: Human Language Technologies-Volume 1}, pages
  987--996. Association for Computational Linguistics.

\bibitem[\protect\citename{Green}2015]{green2015identifying}
Nancy~L Green.
\newblock 2015.
\newblock Identifying argumentation schemes in genetics research articles.
\newblock {\em Proceedings of the 2nd Workshop on Argumentation Mining}, pages
  12--21.

\bibitem[\protect\citename{Habernal \bgroup et al.\egroup }2014]{Habernal2014}
Ivan Habernal, Judith Eckle-Kohler, and Iryna Gurevych.
\newblock 2014.
\newblock {Argumentation Mining on the Web from Information Seeking
  Perspective}.
\newblock In {\em Proceedings of the Workshop on Frontiers and Connections
  between Argumentation Theory and Natural Language Processing}, pages 26--39.

\bibitem[\protect\citename{Hasan and Ng}2014]{Hasan2014}
Kazi~Saidul Hasan and Vincent Ng.
\newblock 2014.
\newblock Why are you taking this stance? identifying and classifying reasons
  in ideological debates.
\newblock In {\em Proceedings of the 2014 Conference on Empirical Methods in
  Natural Language Processing}, pages 751--762.

\bibitem[\protect\citename{Hastings}1963]{hastings1963reformulation}
Arthur~C. Hastings.
\newblock 1963.
\newblock {\em A Reformulation of the Modes of Reasoning in Argumentation}.
\newblock {Ph.D.} thesis.

\bibitem[\protect\citename{Houngbo and Mercer}2014]{Houngbo2014}
Hospice Houngbo and Robert Mercer.
\newblock 2014.
\newblock An automated method to build a corpus of rhetorically-classified
  sentences in biomedical texts.
\newblock pages 19--23.

\bibitem[\protect\citename{Ji and Eisenstein}2015]{Ji2015}
Yangfeng Ji and Jacob Eisenstein.
\newblock 2015.
\newblock One vector is not enough : Entity-augmented distributed semantics for
  discourse relations.
\newblock {\em Transactions of the Association for Computational Linguistics},
  pages 329--344.

\bibitem[\protect\citename{Levy \bgroup et al.\egroup }2014]{Levy2014}
Ran Levy, Yonatan Bilu, Daniel Hershcovich, Ehud Aharoni, and Noam Slonim.
\newblock 2014.
\newblock Context dependent claim detection.
\newblock In {\em Proceedings of COLING 2014, the 25th International Conference
  on Computational Linguistics}, pages 1489--1500.

\bibitem[\protect\citename{Lippi and Torroni}2015]{Lippi2015}
Marco Lippi and Paolo Torroni.
\newblock 2015.
\newblock Argumentation mining: State of the art and emerging trends.
\newblock {\em Association for Computing Machinery Transactions on Internet
  Technology}, pages 1--25.

\bibitem[\protect\citename{Mann and Thompson}1987]{mann1987rhetorical}
William~C Mann and Sandra~A Thompson.
\newblock 1987.
\newblock Rhetorical structure theory: Description and construction of text
  structures.
\newblock In {\em Natural Language Generation}, pages 85--95.

\bibitem[\protect\citename{Peldszus and Stede}2015a]{peldszus2015annotated}
Andreas Peldszus and Manfred Stede.
\newblock 2015a.
\newblock An annotated corpus of argumentative microtexts.
\newblock In {\em Proceedings of the First Conference on Argumentation}, pages
  801--815.

\bibitem[\protect\citename{Peldszus and Stede}2015b]{Peldszus2015}
Andreas Peldszus and Manfred Stede.
\newblock 2015b.
\newblock Joint prediction in mst-style discourse parsing for argumentation
  mining.
\newblock In {\em Proceedings of the 2015 Conference on Empirical Methods in
  Natural Language Processing}, pages 938--948.

\bibitem[\protect\citename{Perelman}1971]{perelman1971new}
Chaim Perelman.
\newblock 1971.
\newblock The new rhetoric.
\newblock In {\em Pragmatics of natural languages}, pages 145--149. Springer.

\bibitem[\protect\citename{Persing and Ng}2016]{Persing2016}
Isaac Persing and Vincent Ng.
\newblock 2016.
\newblock {Modeling Stance in Student Essays}.
\newblock In {\em Proceedings of the 54th Annual Meeting of the Association for
  Computational Linguistics}, pages 2174--2184.

\bibitem[\protect\citename{Prasad \bgroup et al.\egroup }2008]{prasad2008penn}
Rashmi Prasad, Nikhil Dinesh, Alan Lee, Eleni Miltsakaki, Livio Robaldo,
  Aravind~K Joshi, and Bonnie~L Webber.
\newblock 2008.
\newblock The penn discourse treebank 2.0.
\newblock In {\em Proceedings of the 6th Conference on Language Resources and
  Evaluation}.

\bibitem[\protect\citename{Radev}2000]{radev2000common}
Dragomir~R Radev.
\newblock 2000.
\newblock A common theory of information fusion from multiple text sources step
  one: cross-document structure.
\newblock In {\em Proceedings of the 1st SIGdial Workshop on Discourse and
  Dialogue}, pages 74--83.

\bibitem[\protect\citename{Reed \bgroup et al.\egroup }2008]{Reed2008}
Chris Reed, Raquel Mochales~Palau, Glenn Rowe, and Marie-Francine Moens.
\newblock 2008.
\newblock Language resources for studying argument.
\newblock In {\em Proceedings of the 6th Conference on Language Resources and
  Evaluation}, pages 91--100.

\bibitem[\protect\citename{Reed \bgroup et al.\egroup }2016]{reedacl2016}
Chris Reed, Iryna Gurevych, Benno Stein, and Noam Slonim.
\newblock 2016.
\newblock Nlp approaches to computational argumentation.
\newblock In {\em Association for Computational Linguistics (Tutorial)}.

\bibitem[\protect\citename{Reed}2005]{Reed2005}
C~Reed.
\newblock 2005.
\newblock Preliminary results from an argument corpus.
\newblock In {\em The IX Symposium on Social Communication}, pages 576--580.

\bibitem[\protect\citename{Rinott \bgroup et al.\egroup }2015]{rinott2015show}
Ruty Rinott, Lena Dankin, Carlos Alzate, Mitesh~M Khapra, Ehud Aharoni, and
  Noam Slonim.
\newblock 2015.
\newblock Show me your evidence--an automatic method for context dependent
  evidence detection.
\newblock In {\em Proceedings of the 2015 Conference on Empirical Methods in
  Natural Language Processing}, pages 17--21.

\bibitem[\protect\citename{Song \bgroup et al.\egroup }2014]{Song2014}
Yi~Song, Michael Heilman, Beata {Beigman Klebanov}, and Paul Deane.
\newblock 2014.
\newblock Applying argumentation schemes for essay scoring.
\newblock {\em Proceedings of the First Workshop on Argumentation Mining},
  pages 69--78.

\bibitem[\protect\citename{Stab and Gurevych}2014]{Stab2014}
Christian Stab and Iryna Gurevych.
\newblock 2014.
\newblock Annotating argument components and relations in persuasive essays.
\newblock In {\em Proceedings of COLING 2014, the 25th International Conference
  on Computational Linguistics}, pages 1501--1510.

\bibitem[\protect\citename{Stenetorp \bgroup et al.\egroup
  }2012]{stenetorp2012brat}
Pontus Stenetorp, Sampo Pyysalo, Goran Topi{\'c}, Tomoko Ohta, Sophia
  Ananiadou, and Jun'ichi Tsujii.
\newblock 2012.
\newblock Brat: a web-based tool for nlp-assisted text annotation.
\newblock In {\em Proceedings of the Demonstrations at the 13th Conference of
  the European Chapter of the Association for Computational Linguistics}, pages
  102--107.

\bibitem[\protect\citename{Teufel \bgroup et al.\egroup
  }2009]{teufel2009towards}
Simone Teufel, Advaith Siddharthan, and Colin Batchelor.
\newblock 2009.
\newblock Towards discipline-independent argumentative zoning: Evidence from
  chemistry and computational linguistics.
\newblock In {\em Proceedings of the 2009 Conference on Empirical Methods in
  Natural Language Processing}, pages 1493--1502.

\bibitem[\protect\citename{Wachsmuth \bgroup et al.\egroup
  }2016]{wachsmuthusing}
Henning Wachsmuth, Khalid Al-Khatib, and Benno Stein.
\newblock 2016.
\newblock Using argument mining to assess the argumentation quality of essays.
\newblock In {\em Proceedings of COLING 2016, the 26th International Conference
  on Computational Linguistics}, pages 1680--1691.

\bibitem[\protect\citename{Walton \bgroup et al.\egroup
  }2008]{walton2008argumentation}
Douglas Walton, Christopher Reed, and Fabrizio Macagno.
\newblock 2008.
\newblock {\em Argumentation schemes}.
\newblock Cambridge University Press.

\end{thebibliography}

\end{document}